\documentclass[letterpaper]{article} 
\usepackage{aaai25}  
\usepackage{times}  
\usepackage{helvet}  
\usepackage{courier}  
\usepackage[hyphens]{url}  
\usepackage{graphicx} 
\usepackage{natbib}  
\usepackage{caption} 
\usepackage{algorithm}
\usepackage{amsmath,amsthm,amssymb,amsfonts}
\usepackage{color}
\usepackage{graphicx} 
\usepackage{algpseudocodex}
\usepackage{booktabs}
\usepackage{bm}
\usepackage{subcaption}
\usepackage{newfloat}
\usepackage{listings}
\DeclareCaptionStyle{ruled}{labelfont=normalfont,labelsep=colon,strut=off} 
\floatstyle{ruled}
\newfloat{listing}{tb}{lst}{}
\floatname{listing}{Listing}
\title{Mitigating Degree Bias in Signed Graph Neural Networks}
\author {
    Fang He\textsuperscript{\rm 1},
    Jinhai Deng\textsuperscript{\rm 1},
    Ruizhan Xue\textsuperscript{\rm 1},
    Maojun Wang\textsuperscript{\rm 1},
    Zeyu Zhang\textsuperscript{\rm 1}
}
\affiliations {
    \textsuperscript{\rm 1}Huazhong Agricultural University\\

    \{fanghe, dengjinhai, ruizhan\_xue\}@webmail.hzau.edu.cn,
    
    \{mjwang, zhangzeyu\}@mail.hzau.edu.cn
}

\usepackage{bibentry}

\begin{document}

\maketitle

\begin{abstract}
Like Graph Neural Networks (GNNs), Signed Graph Neural Networks (SGNNs) are also up against fairness issues from source data and typical aggregation method. In this paper, we are pioneering to make the investigation of fairness in SGNNs expanded from GNNs. We identify the issue of degree bias within signed graphs, offering a new perspective on the fairness issues related to SGNNs. To handle the confronted bias issue, inspired by previous work on degree bias, a new Model-Agnostic method is consequently proposed to enhance representation of nodes with different degrees, which named as Degree Debiased Signed Graph Neural Network (DD-SGNN) . More specifically, in each layer, we make a transfer from nodes with high degree to nodes with low degree inside a head-to-tail triplet, which to supplement the underlying domain missing structure of the tail nodes and meanwhile maintain the positive and negative semantics specified by balance theory in signed graphs. We make extensive experiments on four real-world datasets. The result verifies the validity of the model, that is, our model mitigates the degree bias issue without compromising performance(\textit{i.e.}, AUC, F1). The code is provided in supplementary material.
\end{abstract}

%

\section{Introduction}\label{Introduction}
Over the past decade, the advent of graph machine learning, particularly Graph Neural Networks (GNNs), has garnered significant attention. GNNs commonly utilize a message passing mechanism, where each node on the graph aggregates data from its neighbors and  produce comprehensive node representations. Just as traditional GNNs, the mainstream Signed Graph Neural Networks (SGNNs) also adopt neighborhood aggregation based on the balance theory \cite{derr2018signed,li2020learning,shu2021sgcl}. It is not difficult to figure out that neighborhood aggregation is the core of all GNNs. Whether for unsigned graphs or signed graphs, GNN-backbones often face fairness and bias issue due to their homogeneity caused by the aggregation mechanisms \cite{chen2022graph, hussain2022adversarial, zhang2023interaction, chen2024fairness, liu2023generalized}. Even within the usage of balance theory, which sets forth two paths—namely, the balanced and the unbalanced—capturing the interplay of both positive and negative relationships, aggregation is essential. That is, in both scenarios, the aggregation process plays a pivotal role, as it coherently integrates these connections and significantly impacts the resulting node representations. 

\begin{figure}
    \centering
    \includegraphics[width=1\linewidth]{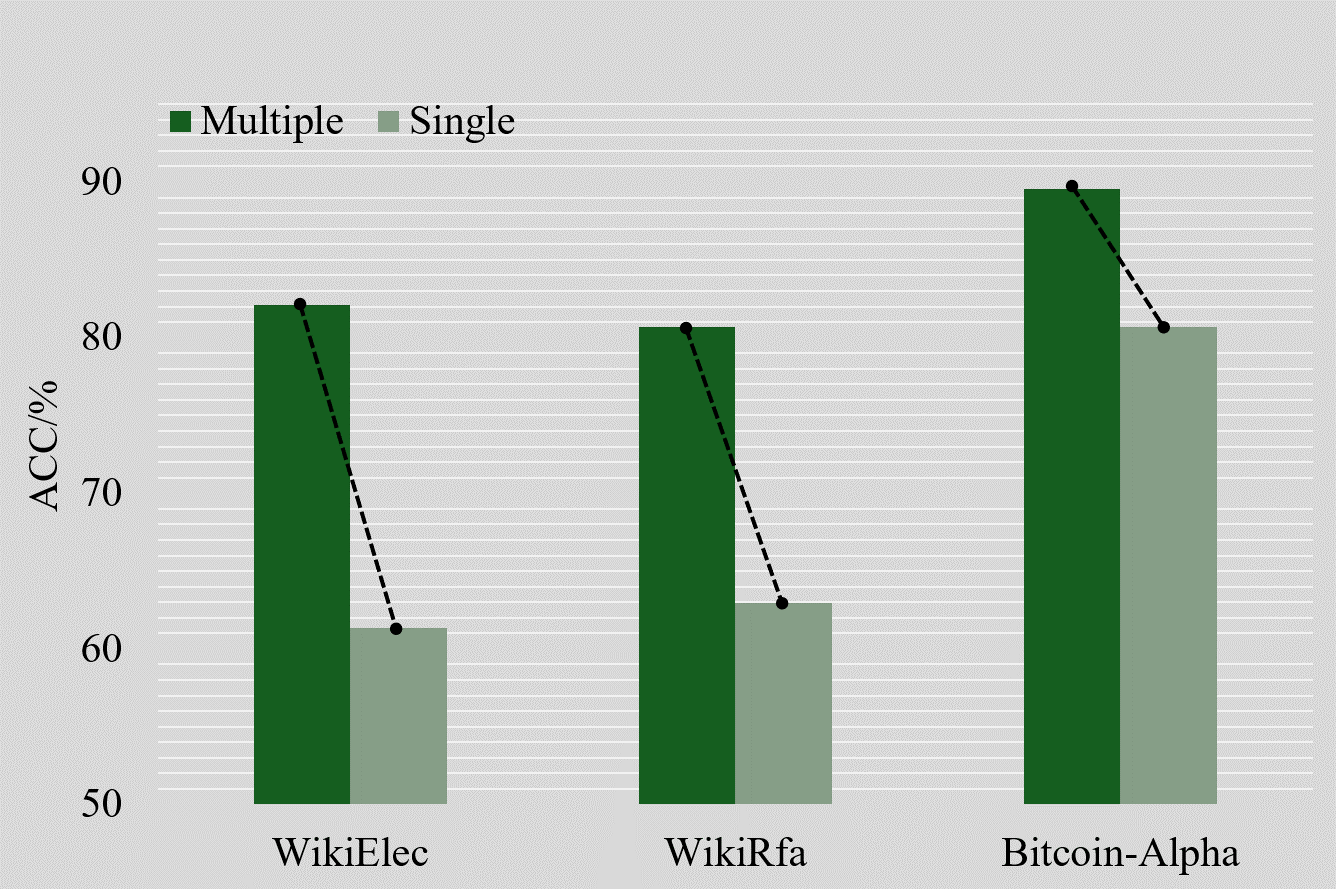}
    \caption{Motivation verification on three real-world datasets. Compare downstream task in accuracy metric between edges connected by `Multiple'-degree nodes and `Single'-degree nodes across signed graph, using the SGCN as the backbone framework, revealing a significant disparity.}
    \label{figure:1}
\end{figure}

Reminding an awareness on this commonality, we further explore whether SGNNs can be confronted with the same fairness issues as GNNs, which showcase in Figure \ref{figure:1}. The result reveals a significant disparity, highlighting the substantial gap in performance between the two groups. In fact, it is not difficult to find the degree-related fairness phenomenon in real world, a polarized landscape of social media can be observed as a clear example of such disparities in action. A polarized signed graph is often a conflicting communities where nodes within each community from dense positive connections and nodes across different communities are sparsely connected via negative links \cite{bonchi2019discovering, huang2022pole}. Inside a community, individuals often exhibit a consensus in their viewpoints. For instance, the relationships among politicians in the U.S. Congress \cite{thomas2006get} and the interactions between Twitter users on political issues \cite{lai2018stance} are representative examples of this phenomenon. The aggregation mechanism makes it easier to predict the edge sign of high degree nodes within polarized communities and harder to do the same for low degree nodes which have a link with another node in different communities, which can be investigated as echo chambers \cite{del2015echo, garrett2009echo, huang2022pole}. These observations and research phenomena reveal a similar \textit{degree bias} issue as it in GNNs.

To address the issue of degree bias, Liu et al. \cite{liu2023fair} explored generalized degree bias in GNNs, and handled the fairness issue in terms of aggregation and degree structure bias. The debiased works in GNNs only accounted for positive interactions such as friendships, acceptances, trust, and support. However, the relationships in real-world are far more complex, often incorporating negative connections that are equally significant with positive connections. `Negative' meanings include adversarial relationships, rejections, distrust, and opposition, which are just as critical to the comprehensive understanding of network dynamics. Having negative links messes up the usual way information is passed, makes the neighborhood structure more complicated. Nodes with high degree are might rich in positive neighborhoods or in negative neighborhoods, we cannot simply employ a positive sampling or a negative sampling to mitigate the degree bias. Hence, it is natural to consider a more rational definition for signed graphs in degree fairness. 

To summarize, we are confronted with two challenges: 
\begin{itemize}
 \item \textbf{Scarce neighborhood on node polarity.} \quad There are many cases in signed graph in terms of the scarcity of node degree. For a node, it might have few positive / negative edges, which makes it lack the information on positive / negative side. A sensible approach is to identify a practical method or context to portray the  scarce neighborhood which is the most relevant standard of node degree, \textit{i.e.}, how to define the degree bias in signed graph? 
 \item \textbf{Multi-Aggregation framework further leads a missing of negative information.} \quad In the first layer, nodes with high degree receive more information from neighborhood than the low ones. In the subsequent layers, different cases (\textit{e.g.}, a two-hop case incorporates high-high, high-low, low-low assemblies) can receive neighborhood information with different richness. The aggregation mechanism in SGNNs was extended to two paths, positive side and negative side, and there are many positive edges and few negative edges in the signed graphs, thus leads to more unfair representations of low degree nodes in negative side. 
\end{itemize}

Based on the motivations and challenges above, to handle \textit{the first issue}, we initially propose a definition of degree debiased within the context of a signed graph, aiming to quantify the degree bias in terms of positive and negative semantics. To define and measure parity within the conventional task of link sign prediction in SGNNs, our objective is essentially related to the final representations of the two nodes engaged in a triplet. Therefore, the pursuit of fairness should commence directly from the examination of these triplets. More specifically, we employ a discriminator determine which triplet the predicted edge belongs to. This approach ensures that the evaluation of fairness is grounded in the actual structure and dynamics of the graph, leading to a more refined and effective mitigation of degree bias. Further more, to handle \textit{the second issue} and to produce better node representations within a \textit{head-to-tail} triplet, we employ a Model-Agnostic framework \textbf{De}gree \textbf{De}biasd \textbf{SGNN}, which supplements missing structural information for the tail node through head-to-tail transferring. Our model operates directly during the first and subsequent aggregation layers of SGNNs and functions as an adaptable plugin, which is designed to bolster the representation of tail nodes and foster a more fair distribution of degrees. The model is generic and can work with any neighbor aggregation based SGNN architecture.

To evaluate the DD-SGNN modle, we employ general AUC and F1 norm to examine and guarantee the performance does not degrade after applying the plugin, together with a fairness-measuring issue to compare basic SGNNs and plugin-added ones under the scale of degree bias context. Extensive experimental evidence demonstrates that our approach is adept at mitigating the fairness issue on the real world datasets without degrading the performance of three based model.

To summarize, our contributions are  as follows.
\begin{itemize}
 \item We are the first to attempt on exploring and addressing degree bias in SGNNs, both definition and measurement are illustrated in the later section.
 \item We propose a Model-Agnostic framework named Degree Debiased SGNN which is a generic plugin and can contribute to more fair representations of nodes with different degrees.
 \item The model is employed in many real-world datasets and mass experiments show that our proposed method can achieve encouraging fair effect with performance comparable to state-of-the-art models.
\end{itemize}

\section{Related Work}\label{Related Work}
\subsubsection{Signed Graph Neural Networks.}\label{Signed Graph Neural Networks.}
The widespread use of social media has made signed networks ubiquitous, sparking interest in their network representations \cite{chen2018bridge, wang2018shine, zhang2023contrastive, zhang2023rsgnn}. While link sign prediction dominates current research, tasks like node classification \cite{tang2016node}, ranking \cite{jung2016personalized}, and community detection \cite{bonchi2019discovering} remain underexplored. Traditional embedding methods like SNE \cite{yuan2017sne}, SIDE \cite{kim2018side}, SGDN \cite{jung2020signed}, and ROSE \cite{javari2020rose} rely on random walks and linear probabilities, but may miss deep relationships. Neural networks, particularly GCN-based SGNNs like SGCN \cite{derr2018signed} and GS-GNN, and GAT-based models like SiGAT \cite{huang2019signed}, SNEA \cite{li2020learning}, SDGNN \cite{huang2021sdgnn}, and SGCL \cite{shu2021sgcl}, are now being applied to capture these complexities, enhancing signed graph representation learning.

\subsubsection{Fairness and Bias in Graph Neural Networks.}\label{Fairness and Bias in Graph Neural Networks.}
Graph neural networks (GNNs)  \cite{kipf2016semi,hamilton2017inductive,velivckovic2017graph} have gained widespread use in the representation learning of nodes/edges/graphs. Despite the great success, recent studies have highlighted that GNNs could introduce various biases from the perspectives of node features and graph topology. The fairness issues related to node features mostly revolve around sensitive attribute values (\textit{e.g.}, gender, age or race)  \cite{chen2024fairness} . The raw features of nodes could be statistically correlated to the sensitive attribute, and thus lead to sensitive information leakage in encoded representations.  Jiang et al.\cite{jiang2022fmp} demonstrating that the message-passing scheme could amplify sensitive node attribute bias. GNNs heavily rely on neighborhood information \cite{wu2020comprehensive} , the trained models and the learned embedding are tied to the neighborhoods, which inadvertently prompts GNNs to perpetuate or amplify biases and discrimination against certain sensitive groups. These indicates that GNNs are largely designed to leverage such bias and unfairness in the data to achieve superior accuracy \cite{dong2023reliant,kose2022fair,liu2023fair,singer2022eqgnn,wang2023fair}. As an example, in terms of the underlying graph structure G used for training is often biased, nodes are more likely to show significant homophily among its neighbors. It may lead to an unfair result that a GNN-based network would have a favor to some specific groups. Despite the fairness of node features, the graph topogy is also a key factor leads to GNN bias issues. Han et al. \cite{han2024towards}  uncover the label position bias indicates that nodes closer to the labeled nodes tends to perform better. Tang et al. \cite{tang2020investigating}  investigated the degree bias in GNNs, signifying that high-degree nodes typically outperform low-degree nodes. Liu et al. \cite{liu2023fair}  define an generalized degree bias to quantificate different multi-hop structures around different nodes. Liu's another research \cite{liu2021tail} also shows the degree bias in graphs, especially the low-degree nodes or so-called tail nodes are likely to be biased or underrepresented. Given these insights, we conducted further investigations to determine whether degree bias is also present in SGCNs, and our  findings validate this phenomenon. 

\section{Methodology}\label{Methodology}
In this section, we will introduce some definitions of related keyword and elaborate on the details of our proposed Model-Agnostic method.

\subsection{Preliminaries}\label{Preliminaries}
\subsubsection{Notations.}\label{Notations.}
Let $\mathcal{G}=(\mathcal{V}, \mathcal{E^{+} },\mathcal{E^{-} })$ be a signed network, where $\mathcal{V}=(v_{1} ,v_{2} ...v_{n} )$ represents the set of n nodes while $\mathcal{E^{+}}\subset\mathcal{V}\times\mathcal{V}$ and $\mathcal{E^{-}}\subset\mathcal{V}\times\mathcal{V}$ denote the sets of positive and negative links, respectively. Note that $\mathcal{E}^{+}\cap\mathcal{E}^{-} = \oslash$, in other words, it is impossible for a pair of nodes to have both positive and negative links simultaneously. Besides $d$ denotes the embedded dimension and $\mathrm {X}^{(0)}\in\mathbb{R}^ {n\times d}$ denotes initial representations.

For every node $v\in \mathcal{V}$, let $\mathcal{N}_{v}$ denote the set of neighboring nodes of $v$, and $\mathcal{N}_{v}^{+}$ means the positive neighborhood, $\mathcal{N}_{v}^{-}$ means the negative ones. 

\subsubsection{Task Definition.} \label{Task Definition.}
In this paper, we address the benchmark task similarly to previous work by \cite{derr2018signed} , \textit{i.e}, the classical link sign prediction task. Formally, given a link $(v_{i},v_{j})$, our goal is to predict the polarity based on final embedding of $v_{i}$ and $v_{j}$. In our work, we disregard edge directions and weights. The link prediction performance can be measured by aligning the ground truth sign and the predicted ones. In terms of fairness, our objective is to estimate the prediction accuracy among three cases, which is discussed in the following section.

\begin{figure*}[t]
\centering
\includegraphics[width=1\textwidth]{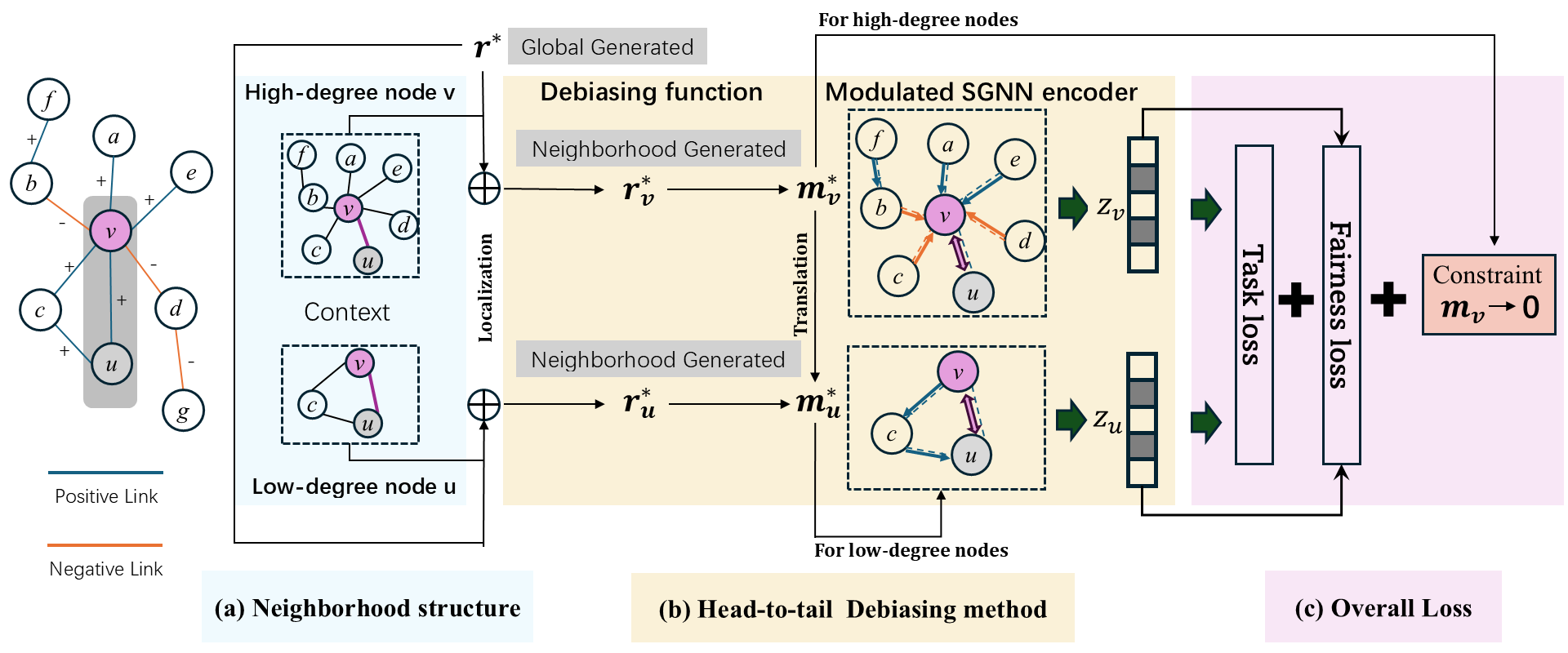} 
\caption{The overall framework of DD-SGNN.}
\label{figure:2}
\end{figure*}

\subsection{Problem Formulation} \label{Problem Formulation}
In this subsection, we will further elaborate on the \textbf{first challenge} and \textbf{motivation} we presented earlier of addressing degree bias on sign graph.

\subsubsection{Degree Bias in SGNN.}
Unlike unsigned graphs, signed graphs feature edges that are categorized into two distinct types: positive and negative. The negative links, as demonstrated in \cite{derr2018signed} , operate on different principles and convey significantly different meanings semantically. Regarding a central node, several scenarios may arise: either an abundance of both positive and negative links, or a situation where the positive links exhibit diversity while the negative links are comparatively sparse. But whether positive or negative information is coherently excavated by balance theory. Consequently, we cannot simplistically sampling head and tail nodes based solely on the presence of positive or negative subgraphs. Instead, a more reasonable perspective is proposed: no matter the dominant of positive or negative links at a head node, we anticipate that it should enrich the local context information of the tail node within a single hop, and extend this supplementary capability over multiple hops to align with the semantics of balance theory. This enrichment aims to mitigate the bias in the tail node's representation during each layer's given aggregation process in SGNN. Hence, we can achieve a more equitable topological structure within signed graphs and ensure fairer node representations for nodes of varying degrees. Specifically, the degree bias on a node stems is defined as $deg(v) = \left|\mathcal{N}_{v}\right|$. As a natural result, we can distinguish the head nodes and the tail nodes by their degree cardinal number $deg(v)$. For some threshold \textit{K}, \textit{tail} nodes are defined as nodes with a degree not exceeding \textit{K}. The tail nodes set $\mathcal{S}_{t}$ is defined as $\mathcal{S}_{t} = \left \{ v\in \mathcal{V} \mid deg(v)\le K \right \}$, whereas \textit{head} nodes set $\mathcal{S}_{h}$ are the complement of tail nodes, \textit{i.e.},  $\mathcal{S}_{h} = \left \{ v\in \mathcal{V} \mid deg(v) > K \right \}$. 

\subsubsection{How to Define a Fairness Issue in SGNN?}
In the context of link prediction, we innovatively formalize a debiasing definition according to statistical parity \cite{dwork2012fairness} . For an edge $e\in \mathcal{E}^{+} \cup \mathcal{E}^{-}$, there is a unique triplet $(v_{i},v_{j},e)$ to confirm $e$. Whereas node $v_{i}$ and $v_{j}$ can be made a head and a tail, or both head, or both tail,
\begin{align*}
 \mathcal{T}_{ht} = \left \{ (v_{i}\in \mathcal{S}_{h}, v_{j}\in \mathcal{S}_{t}),(v_{i}\in \mathcal{S}_{t}, v_{j}\in \mathcal{S}_{h})\mid (v_{i},v_{j},e) \right \},
 \end{align*}
 \begin{align}
 \mathcal{T}_{hh} = \left \{ v_{i},v_{j}\in \mathcal{S}_{h}\mid (v_{i},v_{j},e) \right \},
 \end{align}
 \begin{align*}
 \mathcal{T}_{tt} = \left \{ v_{i},v_{j}\in \mathcal{S}_{t}\mid (v_{i},v_{j},e) \right \}.
 \end{align*}
 To achieve \textit{Degree Statistical Parity} (DSP), we hope whether which triplet the edge in, as Eq.(1) denoting, the prediction of its polarity would be equal:
\begin{equation}
\begin{aligned}
P(\hat{e}=e\mid e\in \mathcal{T}_{hh}) = &P(\hat{e}=e\mid e\in \mathcal{T}_{ht}) \\=&P(\hat{e}=e\mid e\in \mathcal{T}_{tt}),
\end{aligned}
\end{equation}
where $\hat{e}$ is the predicted polarity of $e$.

Specifically, addressing degree bias is to mitigating the significant differences in contextual structures across nodes. This effort aims to achieve fairness and attain a structurally debiased state, facilitating the generation of fair node representations and the accuracy of link prediction. 

\subsection{Implementations of DD-SGNN}
In this section, we introduce the proposed DD-SGNN and address the \textbf{second challenge} we came up with earlier, some of the symbols were given in \textbf{Preliminaries}.

\subsubsection{Aggregation and Propagation in SGNN.} Based on balance theory, SGNNs always maintain two representations at each layer \cite{derr2018signed, li2020learning} . The initial representation of node $v$ is defined as $\mathrm {h}_{v}^{0} \in \mathbb{R}^{d_{in}}$. 
 
\subsubsection{Head-to-Tail Debiasing Method.} 
Inspired by the work of previous scholars \cite{liu2021tail} , the debiasing method from head to tail is crafted to create a tailored debiasing context for every node across all layers, thereby fine-tuning the neighborhood aggregation process (whether in positive aggregation or negative aggregation) on both per-node and per-layer basis. For every node, the debiasing context should encompass both the positive and negative structural information within its neighborhood.

To capture the essence of context embedding within our model, we employ a global shared $\mathrm {r}^{l,*}\in \mathbb{R}^{d_{l}}$ in each layer, $*$ is $pos$ or $neg$ depending on the polarity of links. For the embedding vector $\mathrm {h} _{v} $ of a head node, we define the positive and negative embedding vectors of its neighborhood, denoted by $\mathrm {h}_{\mathcal{N}_{v}^{+}}$ and $\mathrm {h}_{\mathcal{N}_{v}^{-}}$, respectively. Formally, the neighborhood translation on node $v$ in each layer is formulated as 
\begin{equation}
\begin{aligned}
&\mathrm {h}_{v} + \mathrm {r}_{v}^{pos} \approx \mathrm {h}_{\mathcal{N}_{v}^{+}}, 
\\&\mathrm {h}_{v} + \mathrm {r}_{v}^{neg} \approx \mathrm {h}_{\mathcal{N}_{v}^{-}}.
\end{aligned}
\end{equation}
 While $\mathrm {r}_{v}^{pos}$ and $\mathrm {r}_{v}^{neg}$ from head nodes is complete and representative enough, which aims to modeling the local neighborhood information and complement the similar tail nodes within one hop. However, it is essential to recognize that the concept of \textit{missing information} extends beyond a mere absence of naive structural context on either the positive or negative side. Instead, due to the $r^*$ of each layer is generalized and can be expanded by multiple hops in balance theory, it involves a coherent combination that captures implicit semantic information related to missing connections—a representation of `friends' or `enemies'—during the aggregation process. 
 Hence, for $v$ in set $\mathcal{S}_{h}$,
\begin{equation}
\begin{aligned}
     &\mathrm {h}_{v} + \mathrm {r}_{v}^{pos} - \mathrm {h}_{\mathcal{N}_{v}^{+}} \to 0,
     \\&\mathrm {h}_{v} + \mathrm {r}_{v}^{neg} - \mathrm {h}_{\mathcal{N}_{v}^{-}} \to 0.
\end{aligned}
\end{equation}
 On the contrary, the ground truth neighborhood for a tail node is less intact or abundant. The missing neighborhood information of node $v$ in set $\mathcal{S}_{t}$ is defined and predicted as:
\begin{equation}
\begin{aligned}
&\mathrm {m}_{v}^{l,pos}=\mathrm {h}_{v}^l+\mathrm {r}_{v}^{l,pos}-\mathrm {h}_{\mathcal{N}_{v}^{+}}^l,
\\&\mathrm {m}_{v}^{l,neg}=\mathrm {h}_{v}^l+\mathrm {r}_{v}^{l,neg}-\mathrm {h}_{\mathcal{N}_{v}^{-}}^l,
\end{aligned}
\end{equation}
the $\mathrm{r}_{v}^{l,*}$ is made learnable for the predicting how to construct the ideal neighborhood by head nodes, thus transferring knowledge to tail nodes which demonstrate homogeneity to the head ones. 

To reflect the local context of each node, the $\mathrm{r}_{v}^{l}$ is localized by scaling and shifting transformations. More specifically, in the \textit{l}-th layer, given  $\mathrm {r}^{pos}$ and $ \mathrm {r}^{neg}$ in that layer, which can be transformed by a personalization function $\phi$:
\begin{equation}
\begin{aligned}
&\mathrm {r}_{v}^{l,pos} = \phi^{pos}  (\mathrm {h}_{v}^{l},\mathrm {h}_{\mathcal{N}_{v}^{+}}^{l},\mathrm {r}^{pos};\theta _{\phi^{pos}}^{l} ),
\\&\mathrm {r}_{v}^{l,neg} = \phi^{neg}  (\mathrm {h}_{v}^{l},\mathrm {h}_{\mathcal{N}_{v}^{-}}^{l},\mathrm {r}^{neg};\theta _{\phi^{neg}}^{l} ),  
\end{aligned}
\end{equation}
where $\theta _{\phi^{*}}^{l}$ contains the parameters of $\theta$ in the \textit{l}-th layer. Subsequently, we apply scaling and shifting transformations \cite{perez2018film, liu2021tail}. Hence the $\mathrm{r}_{v}^{l}$ can be reformulated as
\begin{equation}
\begin{aligned}
&\phi^{pos}  (\mathrm {h}_{v}^{l},\mathrm {h}_{\mathcal{N}_{v}^{+}}^{l},\mathrm {r}^{pos};\theta _{\phi^{pos}}^{l} ) = (\gamma _{v}^{pos}+\textbf{1})\odot \mathrm {r}^{pos} + \beta _{v}^{pos}, 
\\&\phi^{neg}  (\mathrm {h}_{v}^{l},\mathrm {h}_{\mathcal{N}_{v}^{-}}^{l},\mathrm {r}^{neg};\theta _{\phi^{neg}}^{l} ) = (\gamma _{v}^{neg}+\textbf{1})\odot \mathrm {r}^{neg} + \beta _{v}^{neg},  
\end{aligned}
\end{equation}
$\textbf{1}$ is a vector of ones to ensure the scaling is centered around one.
The $\gamma_{v}^{l} \in \mathbb{R}^{d_{l}}$ is a scaling operator and $\beta_{v}^{l} \in \mathbb{R}^{d_{l}}$ is a shifting operator that are functions of the local context ${\mathrm {h}_{v}^{l},\mathrm {h}_{\mathcal{N}_{v}^{+}}^{l},\mathrm {h}_{\mathcal{N}_{v}^{-}}^{l}}$ in layer \textit{l}, which have the same dimension as the shared translation vector $\mathrm{r}_{v}^{l}$.
\begin{equation}
\begin{aligned}
&\gamma _{v}^{*} = \delta_{\gamma} (\mathrm {W}_{\gamma}^{*,1}\mathrm {h}_{v}^{l}+\mathrm {W}_{\gamma}^{*,2}\mathrm {h}_{\mathcal{N}_{v}^{*}}^{l} ),
\\&\beta_{v}^{*} = \delta_{\beta} (\mathrm {W}_{\beta}^{*,1}\mathrm {h}_{v}^{l}+\mathrm {W}_{\beta}^{*,2}\mathrm {h}_{\mathcal{N}_{v}^{*}}^{l} ),
\end{aligned}
\end{equation}
where $\delta_{\gamma}$ and $\delta_{\beta}$ is
an activation function, and $\mathrm {W}_{*}^{*,1} \in \mathbb{R}^{d_{l}\times{d_{l}}}$ is a learnable weight matrix. Hence, the personalization function $\phi$ is parameterized by these weight matrices:
\begin{equation}
\begin{aligned}
&\theta _{\phi^{pos}}^{l} = \left \{ \mathrm {W}_{\gamma}^{pos,1},\mathrm {W}_{\gamma}^{pos,2},\mathrm {W}_{\beta}^{pos,1},\mathrm {W}_{\beta}^{pos,2} \right \}, 
\\
&\theta _{\phi^{neg}}^{l} = \left \{ \mathrm {W}_{\gamma}^{neg,1},\mathrm {W}_{\gamma}^{neg,2},\mathrm {W}_{\beta}^{neg,1},\mathrm {W}_{\beta}^{neg,2} \right \}.
\end{aligned}
\end{equation}  
Trough transfer strategy from head to tail and predicting assumptions on tail nodes, we operate the $\mathrm{m}_{v}^{l,*}$ in the aggregation process in each layer,  aiming to complete the neighborhood context of tail nodes. 

For the first aggregation layer, the representation is reconstructed as:
\begin{equation}
\begin{aligned}
&\mathbf{h}_{v}^{pos}=\sigma\left({w}^{pos}\left[\left \{\frac {\sum_{j \in \mathcal{N}_{v}^{+}} {\mathbf{h}_{j}^{(0)}}}{\left|\mathcal{N}_{v}^{+}\right|}\right\}\cup\left\{\mathrm{m}_{v}^{pos}\right\},\mathbf{h}_{v}^{(0)}\right]\right),
\\&\mathbf{h}_{v}^{neg}=\sigma\left({w}^{neg}\left[\left \{\frac {\sum_{k \in \mathcal{N}_{v}^{-}} {\mathbf{h}_{k}^{(0)}}}{\left|\mathcal{N}_{v}^{-}\right|}\right\}\cup\left\{\mathrm{m}_{v}^{neg}\right\},\mathbf{h}_{v}^{(0)}\right]\right).
\end{aligned}
\end{equation}

For all subsequent layers (when $l>1$), the aggregation is reconstructed as:
\begin{equation}
\begin{aligned}
\mathbf{h}_{v}^{pos}=\sigma\left({w}^{pos}\left[\left\{\frac{\sum_{j \in \mathcal{N}_{v}^{+}} {\mathbf{h}_{j}^{pos}}}{\left|\mathcal{N}_{v}^{+}\right|} \right \}\cup \left \{ \mathrm {m}_{v}^{pos}\right \},\right. \right.\\
\left.\left.\left\{\frac{\sum_{k \in \mathcal{N}_{v}^{-}} {\mathbf{h}_{k}^{neg}}} {\left|\mathcal{N}_{v}^{-}\right|}\right\}\cup\left\{\mathrm {m}_{v}^{neg}\right \},\mathbf{h}_{v}^{pos}\right]\right),
\\
\mathbf{h}_{v}^{neg}=\sigma\left({w}^{neg}\left[\left\{\frac{\sum_{j \in \mathcal{N}_{v}^{+}} {\mathbf{h}_{j}^{neg}}}{\left|\mathcal{N}_{v}^{+}\right|} \right \}\cup \left \{ \mathrm {m}_{v}^{pos}\right \},\right. \right.\\
\left.\left.\left\{\frac{\sum_{k \in \mathcal{N}_{v}^{-}} {\mathbf{h}_{k}^{pos}}} {\left|\mathcal{N}_{v}^{-}\right|}\right\}\cup\left\{\mathrm {m}_{v}^{neg}\right \},\mathbf{h}_{v}^{neg}\right]\right), 
\end{aligned}
\end{equation}
${w}^{pos}$ and ${w}^{neg}$ is the weight matrices. $\sigma$ is non-linear activation function.

\subsection{Training Constraints and Objective} 
Our task aims to predict the polarity of link $e$ which defined by a unique triplet $(v_{i},v_{j},e)$. We chiefly use the classical signed link prediction loss and fairness loss, together with some other constraints, as shown in Figure \ref{figure:2}, sec.(c).

\subsubsection{Fairness loss.}
According to the fairness issue defined in Eq.(2), the predicted sign of edge $e$ related to final representations of node $v_{i}$ and $v_{j}$ in both degree group (\textit{i.e.},$\mathcal{S}_h$ and $\mathcal{S}_t$) should achieve parity. More specifically, the group $\mathcal{T}_{ht}$ stands an exact head-to-tail transferring method. Thus, in a mind of DSP metric, the fairness loss can be lead as:
\begin{equation}
\begin{aligned}
\mathcal{L}_{1}=\left\|\frac{1}{\left|\mathcal{S}_h\right|}\sum_{\substack{i:\left\{v_i\in\mathcal{S}_h,\right. 
\\ 
\left.\left(v_i,v_j,e\right)\in\mathcal{T}_{ht}\right\}}}\mathbf{z}_i - \frac{1}{\left|\mathcal{S}_t\right|}\sum_{\substack{j:\left\{v_j\in\mathcal{S}_t, \right.
\\
\left.\left(v_i,v_j,e\right)\in\mathcal{T}_{ht}\right\}}}\mathbf{z}_j\right \|_2^2.
\end{aligned}
\end{equation}
It is notable that this loss constrains not a similar final representations but a similar distribution of prediction in our fairness task. More generally, all nodes in a division of head or tail group related to the missing neighborhood information should be constrained. For the head group $\mathcal{S}_h$, its missing neighborhood $\mathrm{m}_{v}^{l,h}=\left\{\mathrm {m}_{v}^{l,pos}\cup \mathrm {m}_{v}^{l,neg}\right\}$ is assumed as flawless, targeting to complement the neighborhood information. For the tail group $\mathcal{S}_t$, its missing neighborhood $m_v^{l,t}$ is predicted by the localized transfer vector from head ones.
The constraint for head nodes promote the learning of debiasing contexts by contrasting the ideal neighborhood, which can be formulated as the following loss.
\begin{equation}
\begin{aligned}
\mathcal{L}_2=  {\textstyle \sum_{l=1}^{L}} \textstyle\sum_{v\in\mathcal{S}_{h}}\left \|\mathrm{m}_{v}^{l,h}\right \|_2^2 ,   
\end{aligned}
\end{equation}
note that each $\left \| \cdot \right \|_2^2 $ beyond refers to the $L_2$ norm.

\subsubsection{Link Sign Prediction Loss.}
For the triplet $(v_{i},v_{j},e)$, we use the final embeddings for node $v_{i}$ and $v_{j}$ (\textit{i.e.}, $[\mathbf{z}_{i}, \mathbf{z}_j]$) as the input to the classifier. Let the concatenated feature vector pass through a MLP layer $\rho $ as a classifier and then a \textit{softmax} function, the main task definition is formalized in the following:
\begin{equation}
\begin{aligned}
& \mathcal{L}_{3}=  -\frac{1}{\left|\mathcal{M}\right|} \sum_{\left(v_i, v_j, e\right) \in \mathcal{M}} \log \frac{\exp \left(\rho_e\left[\mathbf{z}_i, \mathbf{z}_j\right] \right)}{\sum_{s \in\{+,-, ?\}} \exp \left(\rho_s\left[\mathbf{z}_i, \mathbf{z}_j\right] \right)} \\
& +\frac{1}{\left|\mathcal{M}_{(+, ?)}\right|} \sum_{\substack{\left(v_i, v_j, v_k\right) \\
\in \mathcal{M}_{(+, ?)}}} \max \left(0,\left(\left\|\mathbf{z}_i-\mathbf{z}_j\right\|_2^2-\left\|\mathbf{z}_i-\mathbf{z}_k\right\|_2^2\right)\right) \\
& +\frac{1}{\left|\mathcal{M}_{(-, ?)}\right|} \sum_{\substack{\left(v_i, v_j, v_k\right) \\
\in \mathcal{M}_{(-,?)}}} \max \left(0,\left(\left\|\mathbf{z}_i-\mathbf{z}_k\right\|_2^2-\left\|\mathbf{z}_i-\mathbf{z}_j\right\|_2^2\right)\right) \\
& +\operatorname{Reg}\left(\theta^{\mathbf{W}}, \theta^{\rho}\right)
\end{aligned}
\end{equation}
$\mathcal{M}_{(* , ?)}$ are the set for the pairs of positively or negatively linked nodes, $s\in\{+,-,?\}$ represents $s$ was a positive, negative, or no link between nodes pair $(v_{i},v_{j})$ . For every linked node pair in triplets, there is another randomly sampled $v_k$ for achieving our task, and $\operatorname{Reg}\left(\theta^{\mathbf{W}}, \theta^{\rho}\right)$ is the regularization goal.

\subsubsection{Overall loss.}
Finally, taking all the above loss items into account, we can express the total loss as
\begin{equation}
\begin{aligned}
    \mathcal{L}=\mu\mathcal{L}_1 + \eta\mathcal{L}_2+ \mathcal{L}_3,
\end{aligned}
\end{equation}
where $\mu$, $\eta$ are hyper-parameters.  

\section{Experiment}
In this section, we begin by introducing our datasets, and evaluate the proposed DD-SGNN in terms of both accuracy and fairness.
\subsection{Experimental Setup}
\subsubsection{Datasets.}
We conducted experiments on four real-world signed network datasets that reflect various online communities. Bitcoin-Alpha and Bitcoin-OTC \cite{kumar2016edge} are both networks where each node corresponds to a user from the Bitcoin-OTC platform, and the edges illustrate the dynamics of trust and distrust among users.
Wikirfa \cite{west2014exploiting} shifts the focus to Wikipedia's community governance, capturing the collaborative decision-making process through support and opposition votes in adminship elections. Slashdot \cite{kumar2016edge} is from a technology community, recording interactions that show users' likes and dislikes, indicating agreement or disagreement with each other's posts. These datasets are summarized in Table \ref{tab：t1} with more statistics. 

\begin{table}[ht]
\centering
\begin{tabular}{@{}llllc@{}}
\toprule
Dataset       & \#Node & \#Edge & \%Pos & Source       \\ \midrule
Bitcoin-Alpha & 3,783  & 24,186  & 93.7  & Real dataset \\
Bitcoin-OTC   & 5,881  & 35,592  & 90.9  & Real dataset \\
Wikirfa       & 11,259 & 178,096 & 77.9  & Real dataset \\ 
Slashdot      & 82144  & 549202  & 77.4  & Real dataset \\ \bottomrule
\end{tabular}
\caption{Common Datasets for Signed Networks}
\label{tab：t1}
\end{table}

\subsubsection{Base SGNN models.}
DD-SGNN functions to counteract biases magnified through aggregation, thus being model-agnostic and being employed with three base SGNN model. By default, we employ SGCN \cite{derr2018signed} as the base SGNN model in our experiments. Besides, the popular model SNEA \cite{li2020learning} and SGCL\cite{shu2021sgcl} are also set as the base SGNN to evaluate the flexibility of our debiasing function, \textit{i.e.}, DD-SNEA and DD-SGCL, respectively. The comparisons are listed in Table \ref{tab:t2}. These two base models  presents a greater degree bias than SGCN, further confirming aggregation worsens fairness of structure.

\subsubsection{Data split and parameters.}
For each dataset, we randomly split the edges into training and test set with proportion
8:2. We set the threshold \textit{K} for the structural contrast as the mean node degree by default. The number and range of 
the foremost \textit{K} is further explored in Appendix. 

\subsubsection{Evaluation.}
For model \textit{performance}, we utilize both F1 and Area Under the receiver operating characteristic Curve (AUC). Higher F1 and AUC both mean better performance. 

For model \textit{fairness}, the DSP essentially require that the outcomes are independent of the \textit{head-tail} groups, as discussed before. To this end, we have divided the test dataset into three distinct groups: $\mathcal{T}_{hh}$, $\mathcal{T}_{ht}$ and $\mathcal{T}_{tt}$, containing test triplets with degree standard, respectively. Our fairness objective is to achieve greater parity, that is, to lower the DSP score and reduce the disparity in the final link prediction accuracy across nodes with different degrees. Furthermore, real-world datasets often exhibit a scarcity of triplets in the group $\mathcal{T}_{tt}$, given the set threshold \textit{K} within our study. This scarcity implies a lack of representativeness. Hence, the criteria for grouping naturally converge to two categories: both head entities, $\mathcal{T}_{hh}$, and the other that does not differentiate between head and tail, $\mathcal{T}_{\cdot t}$. 
Hence, the metric $\Delta_{\mathrm{DSP}}$, which measures the prediction bias between the two groups based on DSP, is defined as:
\begin{equation}
\begin{aligned}
\hspace{-3.5mm}
\quad \Delta_{\mathrm{DSP}}=\frac{1}{\left |\mathcal{E}\right |} {\sum_{e\in\mathcal{E}}}\left| P\left (\hat{e}=e\mid e\in \mathcal{T}_{hh} \right) - P\left ( \hat{e}=e\mid e\in \mathcal{T}_{\cdot t} \right ) \right |. 
\end{aligned}
\end{equation}
The metric $\Delta_{\mathrm{DSP}}$ and DSP will not  be made in distinction in the following text.

\subsection{Model Performance and Fairness}
We first evaluate our model using the default base SGNN (\textit{i.e.}, SGCN) and fairness settings, and further supplement it with additional fairness settings and base SGNNs.

\subsubsection{Main Link Prediction evaluation.}
We apply the plugin to enhance model performance in three baselines, as comparison shown in Table \ref{tab:t2}, and make the following observations. 

Firstly, DD-SGNN consistently outperforms baseline on fairness target, and maintains a level almost consistent with the baseline on AUC and F1. In typical GNNs, there is often a trade-off between model performance and fairness metric \cite{bose2019compositional, dai2021say,yang2024fairsin}. However, SGNNs fails to clearly exhibit such a tendency. Instead, the magnitude of balance factor is more correlated with original data set. Bitcoin-Alpha has a high probability of $93.7\%$ in positive links, making it less polar and less inclined to a `perfect' signed graph, in which can be observed a relatively lower DSP than other Daraset, \textit{e.g.}, WikiRfa. This phenomenon proves that ``there are many positive edges and few negative edges in the signed graphs, thus lead to more unfair representations of low degree nodes on negative side" we proposed before. That is, if a signed graph fails to incline polar obviously, it is more closer to a unsigned graph and would show less bias, whose view is similarly reflected in other works \cite{huang2022pole}. This allied appearance provides us with the notion that within signed graphs, fairness issue is highly likely to be associated with the polarity of the graph and the polarization phenomenon.  
\begin{table*}[ht]
\resizebox{\textwidth}{!}{
    \setlength{\tabcolsep}{10pt}
    \begin{tabular}{@{}lr||ccc|ccc@{\hspace{10pt}}}
    \toprule
                                       &                                                              & SGCN & SNEA & SGCL & DD-SGCN & DD-SNEA & DD-SGCL \\ \midrule\midrule
    \multicolumn{1}{l|}{Bitcoin-Alpha} &
  \begin{tabular}[c]{@{}r@{}}AUC↑\\ FI ↑\\ DSP↓\end{tabular} &
  \multicolumn{1}{r}{\begin{tabular}[c]{@{}r@{}}83.36±0.31\\ 93.22±0.06\\ 1.17±0.26\end{tabular}} &
  \multicolumn{1}{r}{\begin{tabular}[c]{@{}r@{}}83.61±0.06\\ 92.62±0.11\\ 1.89±0.18\end{tabular}} &
  \multicolumn{1}{r|}{\begin{tabular}[c]{@{}r@{}}83.99±0.17\\ 91.20±0.81\\ 2.12±0.19\end{tabular}} &
  \multicolumn{1}{r}{\begin{tabular}[c]{@{}r@{}}\underline{84.38}±0.97\\ \textbf{93.54}±0.16\\ \textbf{0.46}±0.14\end{tabular}} &
  \multicolumn{1}{r}{\begin{tabular}[c]{@{}r@{}}83.89±0.17\\ \underline{93.33}±0.16\\ \underline{0.69}±0.27\end{tabular}} &
  \multicolumn{1}{r}{\begin{tabular}[c]{@{}r@{}}\textbf{88.03}±1.51\\ 92.54±0.15\\ 1.51±0.23\end{tabular}} \\ \midrule

    \multicolumn{1}{l|}{Bitcoin-OTC} &
  \begin{tabular}[c]{@{}r@{}}AUC↑\\ FI ↑\\ DSP↓\end{tabular} &
  \multicolumn{1}{r}{\begin{tabular}[c]{@{}r@{}}85.77±0.07\\ \textbf{93.76}±0.23\\ \underline{2.53}±0.07\end{tabular}} &
  \multicolumn{1}{r}{\begin{tabular}[c]{@{}r@{}}\underline{86.14}±0.05\\ 92.23±0.34\\ 8.70±0.63\end{tabular}} &
  \multicolumn{1}{r|}{\begin{tabular}[c]{@{}r@{}}71.58±0.54\\ 86.62±1.09\\ 7.42±0.98\end{tabular}} &
  \multicolumn{1}{r}{\begin{tabular}[c]{@{}r@{}}85.06±0.21\\ 93.15±0.45\\ \textbf{1.34}±0.30\end{tabular}} &
  \multicolumn{1}{r}{\begin{tabular}[c]{@{}r@{}}\textbf{86.65}±0.53\\ 92.61±0.29\\ 7.55±1.20\end{tabular}} &
  \multicolumn{1}{r}{\begin{tabular}[c]{@{}r@{}}75.00±1.26\\ \underline{93.55}+1.28\\ 6.86±0.58\end{tabular}} \\ \midrule

    \multicolumn{1}{l|}{SlashDot} &
  \begin{tabular}[c]{@{}r@{}}AUC↑\\ FI ↑\\ DSP↓\end{tabular} &
  \multicolumn{1}{r}{\begin{tabular}[c]{@{}r@{}}\underline{78.33}±0.02\\ 86.45±0.27\\ \underline{7.94}+0.16\end{tabular}} &
  \multicolumn{1}{r}{\begin{tabular}[c]{@{}r@{}}67.25±0.03\\ 81.86±0.30\\ 13.18±1.35\end{tabular}} &
  \multicolumn{1}{r|}{\begin{tabular}[c]{@{}r@{}}75.42±0.59\\ 82.22±0.24\\ 10.25±1.08\end{tabular}} &
  \multicolumn{1}{r}{\begin{tabular}[c]{@{}r@{}}\textbf{78.86}±0.66\\ \textbf{89.06}±0.37\\ \textbf{3.84}±0.97\end{tabular}} &
  \multicolumn{1}{r}{\begin{tabular}[c]{@{}r@{}}65.75±0.05\\ 81.24±0.34\\ 15.32±0.12\end{tabular}} &
  \multicolumn{1}{r}{\begin{tabular}[c]{@{}r@{}}76.88±0.85\\ \underline{87.17}±0.79\\ 8.41±0.88\end{tabular}} \\ \midrule

    \multicolumn{1}{l|}{WikiRfa}       & \begin{tabular}[c]{@{}r@{}}AUC↑\\  FI ↑\\  DSP↓\end{tabular} &
    \multicolumn{1}{r}{\begin{tabular}[c]{@{}r@{}}\underline{79.1}±0.06\\ \underline{86.14}±0.08\\ 5.68±0.46\end{tabular}} &
    \multicolumn{1}{r}{\begin{tabular}[c]{@{}r@{}}75.02±0.02\\ 84.6±0.46\\ 6.41±0.17\end{tabular}} &
    \multicolumn{1}{r|}{\begin{tabular}[c]{@{}r@{}}71.62±0.62\\ 85.13±1.40\\ 7.25±0.13\end{tabular}} &
    \multicolumn{1}{r}{\begin{tabular}[c]{@{}r@{}}\textbf{79.62}±0.43\\ \textbf{86.54}±0.12\\ \textbf{3.70}±0.31\end{tabular}} &
    \multicolumn{1}{r}{\begin{tabular}[c]{@{}r@{}}76.44±0.32\\ 84.05±0.46\\ 6.41±0.17\end{tabular}} &
    \multicolumn{1}{r}{\begin{tabular}[c]{@{}r@{}}76.46±1.18\\ 85.45±0.23\\ \underline{5.03}±0.33\end{tabular}}  \\ \bottomrule
    \end{tabular}
}
\caption{Comparison with baselines ($\mathcal{T}_{hh}$ \&  $\mathcal{T}_{\cdot t}$).}
\subcaption*{Tabular results are in percent; standard deviation more than 5 runs; the best result is \textbf{bolded} and the runner-up is \underline{underlined}.}
\label{tab:t2}
\end{table*}
\subsubsection{Fairness settings.}
We further evaluate fairness $\Delta_{\mathrm{DSP}}$ using different triple groups, \textit{i.e.}, metric between $\mathcal{T}_{hh}$ and $\mathcal{T}_{ht}$ in Table \ref{tab：t3} together with $\mathcal{T}_{ht}$ and $\mathcal{T}_{tt}$ in Table \ref{tab：t4}. Note that, the data being evaluated for these two comparative groups are not fully derived from the test set, instead, distinct threshold values \textit{K} are assigned to the head and tail nodes, tailored to the degree distributions inherent in different original datasets. More specifically, the upper 20th percentile is designated as high-degree head nodes, while the lower 20th percentile is categorized as low-degree tail nodes. Here, we compare SGCN and DD-SGCN, and observe that, even with different group lines, our proposed DD-SGNN can perform better than baseline in terms of degree fairness. Note that, the method especially improves fairness between group $\mathcal{T}_{ht}$ and $\mathcal{T}_{tt}$, indicating that degree bias is effectively alleviated.
\begin{table}[ht]
\centering
\footnotesize
    \begin{tabular}{c@{\hspace{0pt}}r@{\hspace{0pt}}||cc@{\hspace{1pt}}|c@{\hspace{9pt}}}
    \toprule
                                       &                                                              & SGCN   & DD-SGCN & Decline \\ \midrule\midrule
    \multicolumn{1}{l}{Bitcoin-Alpha}       & \begin{tabular}[c]{@{}r@{}}\end{tabular} &
    \multicolumn{1}{r}{\begin{tabular}[c]{@{}r@{}}0.60±0.21\end{tabular}} &
    \multicolumn{1}{r|}{\begin{tabular}[c]{@{}r@{}}\textbf{0.43}±0.25\\\end{tabular}} &
    \multicolumn{1}{r}{\begin{tabular}[c]{@{}r@{}}28.33\%↓\end{tabular}} \\ \midrule

    \multicolumn{1}{l}{Bitcoin-OTC}       & \begin{tabular}[c]{@{}r@{}}\end{tabular} &
    \multicolumn{1}{r}{\begin{tabular}[c]{@{}r@{}}2.49±0.27\end{tabular}} &
    \multicolumn{1}{r|}{\begin{tabular}[c]{@{}r@{}}\textbf{1.60}+0.17\\\end{tabular}} &
    \multicolumn{1}{r}{\begin{tabular}[c]{@{}r@{}}35.74\%↓\end{tabular}} \\ \midrule

    \multicolumn{1}{l}{SlashDot}       & \begin{tabular}[c]{@{}r@{}}\end{tabular} &
    \multicolumn{1}{r}{\begin{tabular}[c]{@{}r@{}}9.40±0.27\end{tabular}} &
    \multicolumn{1}{r|}{\begin{tabular}[c]{@{}r@{}}\textbf{9.30}+0.48\\\end{tabular}} &
    \multicolumn{1}{r}{\begin{tabular}[c]{@{}r@{}}1.06\%↓\end{tabular}} \\ \midrule

    \multicolumn{1}{l}{WikiRfa}       & \begin{tabular}[c]{@{}r@{}}\end{tabular} &
    \multicolumn{1}{r}{\begin{tabular}[c]{@{}r@{}}11.48±0.16\end{tabular}} &
    \multicolumn{1}{r|}{\begin{tabular}[c]{@{}r@{}}\textbf{10.87}±0.53\\\end{tabular}} &
    \multicolumn{1}{r}{\begin{tabular}[c]{@{}r@{}}5.31\%↓\end{tabular}} \\ \bottomrule    
    \end{tabular}
\caption{Comparison on DSP ($\mathcal{T}_{hh}$ \& $\mathcal{T}_{ht}$, 20\% T/B)}
\label{tab：t3}
\end{table}
\begin{table}[ht]
\centering
\footnotesize
    \begin{tabular}{c@{\hspace{0pt}}r@{\hspace{0pt}}||cc@{\hspace{1pt}}|c@{\hspace{9pt}}}
    \toprule
                                       &                                                              & SGCN   & DD-SGCN & 
  Decline \\ \midrule\midrule
    \multicolumn{1}{l}{Bitcoin-Alpha}       & \begin{tabular}[c]{@{}r@{}}\end{tabular} &
    \multicolumn{1}{r}{\begin{tabular}[c]{@{}r@{}}5.06±0.39\end{tabular}} &
    \multicolumn{1}{r|}{\begin{tabular}[c]{@{}r@{}}\textbf{3.59}±0.39\end{tabular}} &
    \multicolumn{1}{r}{\begin{tabular}[c]{@{}r@{}}29.05\%↓\end{tabular}} \\ \midrule

    \multicolumn{1}{l}{Bitcoin-OTC}       & \begin{tabular}[c]{@{}r@{}}\end{tabular} &
    \multicolumn{1}{r}{\begin{tabular}[c]{@{}r@{}}4.52±0.12\end{tabular}} &
    \multicolumn{1}{r|}{\begin{tabular}[c]{@{}r@{}}\textbf{2.36}±0.53\end{tabular}} &
    \multicolumn{1}{r}{\begin{tabular}[c]{@{}r@{}}47.79\%↓\end{tabular}} \\ \midrule

    \multicolumn{1}{l}{SlashDot}       & \begin{tabular}[c]{@{}r@{}}\end{tabular} &
    \multicolumn{1}{r}{\begin{tabular}[c]{@{}r@{}}11.89±0.18\end{tabular}} &
    \multicolumn{1}{r|}{\begin{tabular}[c]{@{}r@{}}\textbf{8.52}±0.29\end{tabular}} &
    \multicolumn{1}{r}{\begin{tabular}[c]{@{}r@{}}28.34\%↓\end{tabular}} \\ \midrule

    \multicolumn{1}{l}{WikiRfa}       & \begin{tabular}[c]{@{}r@{}}\end{tabular} &
    \multicolumn{1}{r}{\begin{tabular}[c]{@{}r@{}}14.36±0.23\end{tabular}} &
    \multicolumn{1}{r|}{\begin{tabular}[c]{@{}r@{}}\textbf{11.35}±0.35\end{tabular}} &
    \multicolumn{1}{r}{\begin{tabular}[c]{@{}r@{}}20.96\%↓\end{tabular}} \\ \bottomrule
    \end{tabular}
\caption{Comparison on DSP ($\mathcal{T}_{ht}$ \& $\mathcal{T}_{tt}$, 20\% T/B)}
\label{tab：t4}
\end{table}
\subsubsection{Ablation Study.}
To ensure that each module in DD-SGNN contributes, we considered several variants. (1)\textit{no missing information and translation}: we remove the module which modeling the missing context by neighborhood translation. (2)\textit{no constraint for head nodes}: we remove the module that ensure the ideal neighborhood information of head nodes. (3)\textit{no localization}: we remove the scale and shift for each node, meaning a lack of node adaption. 

\begin{figure}[ht]
    \centering
    \subsubsection{}
    {\includegraphics[width=0.8\linewidth]{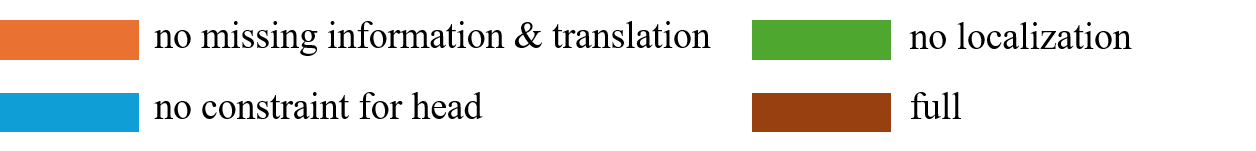}}
    \subcaptionbox{AUC\%↑}{\includegraphics[width=0.3\linewidth]{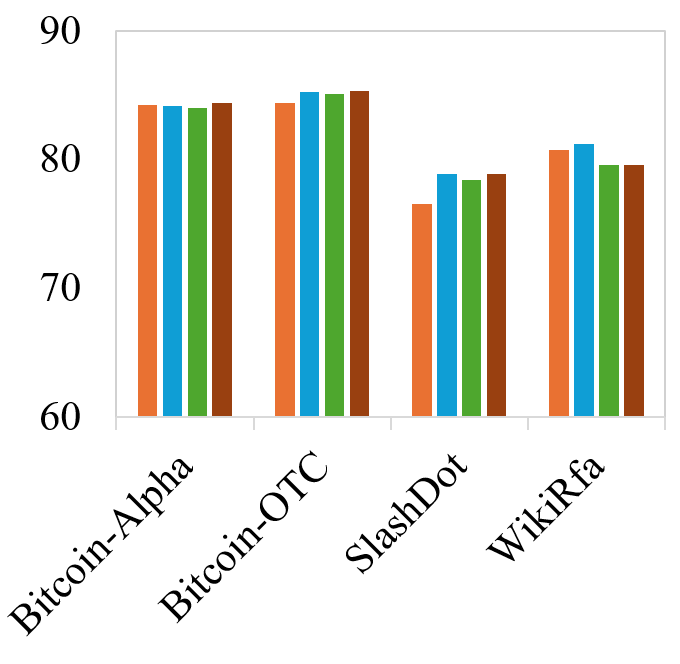}}
    \hfill
    \subcaptionbox{F1\%↑}{\includegraphics[width=0.3\linewidth]{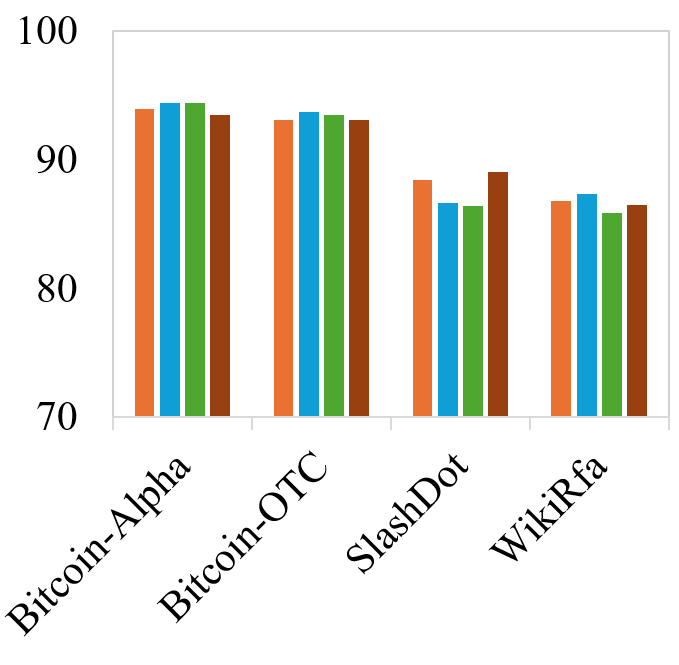}}
    \hfill
    \subcaptionbox{DSP↓}{\includegraphics[width=0.3\linewidth]{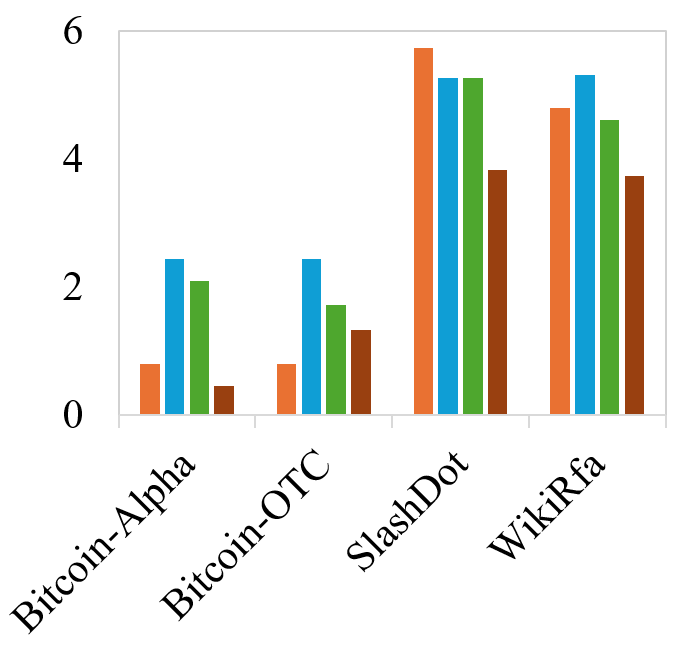}}

    \caption{Ablation study on the effect of each module.}
    \label{figure:3}
\end{figure}

The result is reported in Figure \ref{figure:3}. We can observed that no matter which module is removed, the model shows almost no better performance in fairness metric. After removing the translation technique, which means that transfer cannot be applied, model tends to show particular poor fairness. This indicates that our model gains advantages from this module. Similarly, both the AUC and F1 indicators decreased after the removal of each module.
\subsection{Conclusions}
In this paper, we pioneeringly investigated 
the degree bias on SGNNs. For the first time, we attempt to define and address the degree bias issue within signed graphs. To alleviate the bias in positive and negative information (especially the negative) caused by varying node degrees in multi-layer neighborhood aggregation, we propose a framework called DD-SGNN, which can work in conjunction with various neighborhood aggregations. The key to this framework is to use transfer learning techniques to supplement the positive and negative information of tail nodes within one hop of the head node. We conducted extensive experiments on four real-world datasets and the result on performance and fairness metric of our model were highly satisfactory.

\bibliography{aaai25}
\clearpage
\appendix
\begin{huge}
    \section{Appendix}
\end{huge}
\subsection{Hyper-parameter Discuss}
The hyper-parameter \textit{K} is further experimented in range of [6, 15, 30, 50] in different datasets. Two datasets with high positive side ratio, \textit{i.e.}, Bitcoin-Alpha and Bitcoin-OTC, both shows a low peak on the value of 30. The other two datasets with greater polarity show an obvious downward trend with the increase of \textit{K}.
\begin{figure}[ht]
    \centering
    \includegraphics[width=1\linewidth]{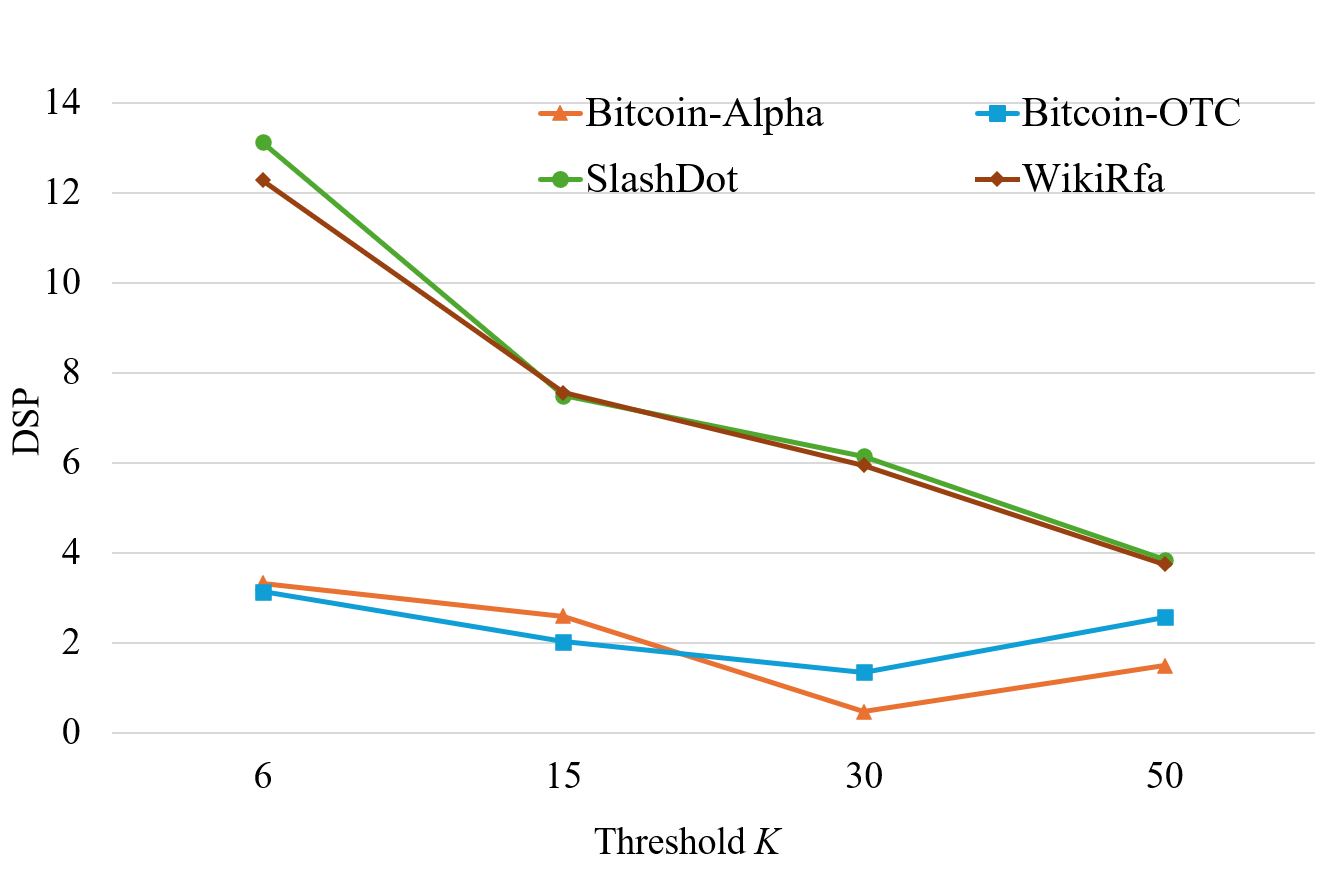}
    \caption{DSP performance under different values of hyper-parameter \textit{K}.}
    \label{figure:4}
\end{figure}
\end{document}